%% file: cnnmodularity.tex
\documentclass[sigconf]{acmart}
\input{macro.tex}

\usepackage{subfig}
\usepackage{hhline}
\usepackage{adjustbox}
\usepackage{diagbox}
\usepackage[justification=centering]{caption}
\definecolor{moccasin}{rgb}{0.98, 0.92, 0.84}
\definecolor{magicmint}{rgb}{0.67, 0.94, 0.82}
\AtBeginDocument{%
	\providecommand\BibTeX{{%
			\normalfont B\kern-0.5em{\scshape i\kern-0.25em b}\kern-0.8em\TeX}}}

\copyrightyear{2021}
\acmYear{2021}
\setcopyright{acmcopyright}\acmConference[ICSE 2022]{The 44th International Conference on Software Engineering}{May 21-29, 2022}{Pittsburgh, PA, USA}
\graphicspath{{./figures/}}

\begin{document}
	
	\title{Decomposing Convolutional Neural Networks into Reusable and Replaceable Modules}

	\author{Rangeet Pan}
	\email{rangeet@iastate.edu}
	\affiliation{%
		\institution{Dept. of Computer Science, Iowa State University}
		\streetaddress{226 Atanasoff Hall}
		\city{226 Atanasoff Hall, Ames}
		\state{IA}
		\postcode{50011}
		\country{USA}
	}

	\author{Hridesh Rajan}
	\email{hridesh@iastate.edu}
	\affiliation{%
		\institution{Dept. of Computer Science, Iowa State University}
		\streetaddress{226 Atanasoff Hall}
		\city{226 Atanasoff Hall, Ames}
		\state{IA}
		\postcode{50010}
		\country{USA}
	}
	
	
	\input{abstract}

	\ccsdesc[500]{Computing methodologies~Machine learning}
	\ccsdesc[500]{Software and its engineering~Abstraction and modularity}
	
	\keywords{deep learning, cnn, deep neural network, modularity, decomposition}
	

	\maketitle
	
	\input{introduction.tex}
	\input{related.tex}

\input{methodology}
	\input{technique1.tex}
	\input{technique3.tex}
	\input{results}
	\input{rq1.tex}
	\input{rq2.tex}
	\input{rq3.tex}
	\input{conclusion.tex}
	\balance
	
	\bibliographystyle{ACM-Reference-Format}
	\bibliography{refs}
	
	
\end{document}

%% file: macro.tex
\usepackage{algorithm}
\usepackage[noend]{algpseudocode}
\usepackage{multirow}
\usepackage{booktabs} 
\usepackage{amsmath}
\usepackage{graphicx}
\usepackage{caption}
\usepackage{rotating}
\usepackage{tabularx}
\usepackage[
font=footnotesize 
]{subfig}
\usepackage{wrapfig}
\usepackage{listings}
\usepackage{color, colortbl}
\usepackage{balance} 
\usepackage{lscape}
\usepackage{hyperref}
\usepackage{xspace}
\usepackage{framed}
\usepackage{syntax}
\usepackage{soul}
\usepackage{flushend}
\usepackage[tikz]{bclogo}
\usepackage{makecell}
\definecolor{codegreen}{rgb}{0,0.6,0}
\definecolor{codegray}{rgb}{0.5,0.5,0.5}
\definecolor{codepurple}{rgb}{0.58,0,0.82}
\definecolor{backcolour}{rgb}{0.95,0.95,0.92}
\definecolor{Gray}{gray}{0.1}
\lstdefinestyle{mystyle}{
	backgroundcolor=\color{backcolour},   
	commentstyle=\color{codegreen},
	keywordstyle=\color{magenta},
	numberstyle=\tiny\color{codegray},
	stringstyle=\color{codepurple},
	basicstyle=\scriptsize,
	breakatwhitespace=false,         
	breaklines=true,                 
	captionpos=b,                    
	keepspaces=true,                 
	numbers=left,                    
	numbersep=5pt,                  
	showspaces=false,                
	showstringspaces=false,
	showtabs=false,                  
	tabsize=2
}

\lstdefinelanguage{Pythonna}{%
	language     = python,
	morekeywords = {to_categorical, flow_from_directory,pad_sequences,load_image},
	morecomment=[f][\color{diffstart}]{@@},
	morecomment=[f][\color{diffincl}]{+\ },
	morecomment=[f][\color{diffrem}]{-\ }	
}

\lstdefinestyle{customc}{
	belowcaptionskip=1\baselineskip,
	breaklines=false,
	frame= single,
	breaklines = true,
	xleftmargin=\parindent,
	language= Pythonna,
	showstringspaces=false,
	basicstyle=\footnotesize\ttfamily,
	keywordstyle=\bfseries\color{green!40!black},
	commentstyle=\itshape\color{purple!40!black},
	identifierstyle=\color{blue},
	stringstyle=\color{codegreen},
	backgroundcolor=\color{gray!4}
}

\graphicspath{{./figures/}}

\newcommand{\etal}{{\em et al.}\xspace}

\newcommand{\conv}{\textit{Convolution}\xspace}

\newcounter{rqs}
\stepcounter{rqs}

\newcounter{NumObservations}
\stepcounter{NumObservations}
\definecolor{shadecolor}{rgb}{.9,.9,.9}

%% file: abstract.tex
\begin{abstract}
	Training from scratch is the most common way to build a Convolutional Neural Network (CNN) based model. 
	What if we can build new CNN models by reusing parts from previously build CNN models? 
	What if we can improve a CNN model by replacing (possibly faulty) parts with other parts?
	In both cases, instead of training, can we identify the part responsible for each output class (module) in the model(s) and reuse or replace only the desired output classes to build a model?
	Prior work has proposed decomposing dense-based networks into modules (one for each output class) to enable reusability and replaceability in various scenarios. However, this work is limited to the dense layers and based on the one-to-one relationship between the nodes in consecutive layers. Due to the shared architecture in the CNN model, prior work cannot be adapted directly. 
	In this paper, we propose to decompose a CNN model used for image classification problems into modules for each output class. These modules can further be reused or replaced to build a new model. 
	We have evaluated our approach with CIFAR-10, CIFAR-100, and ImageNet tiny datasets with three variations of ResNet models and found that enabling decomposition comes with a small cost (1.77\% and 0.85\% for top-1 and top-5 accuracy, respectively). Also, building a model by reusing or replacing modules can be done with a 2.3\% and 0.5\% average loss of accuracy. 
	Furthermore, reusing and replacing these modules reduces \textbf{$CO_2e$} emission by $\sim$37 times compared to training the model from scratch.
\end{abstract}

%% file: introduction.tex
\section{Introduction}
\label{sec:intro}
\begin{figure*}%
	\centering
	\footnotesize
	\subfloat{\includegraphics[, width=1.0\linewidth]{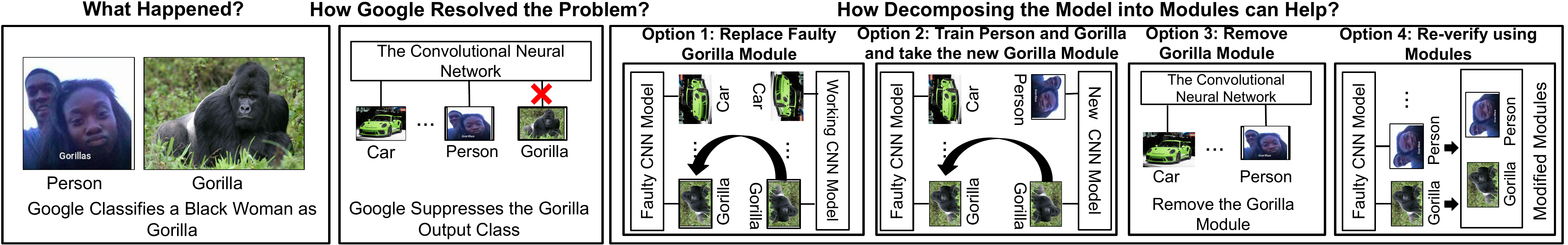}}%
	\qquad
	\subfloat{\input{tbMotiv.tex}}
	\caption{How Decomposing a Convolution Neural Network into Modules Can Help Solve a Real-life Problem.}%
	\label{fig:overview}%
\end{figure*}

Deep learning is increasingly being used for image segmentation, object detection, and similar computer vision tasks. With the need for bigger and complex datasets, the size and complexity of models also increase. Often, training a model from scratch needs several hours or even days. 
To ease the training process, layer architecture~\cite{sun2018resinnet, luo2017thinet}, transfer learning~\cite{torrey2010transfer, tan2018survey}, one-shot learning~\cite{vinyals2016matching, rezende2016one}, few-shot learning~\cite{sung2018learning, sun2019meta, lifchitz2019dense, wertheimer2019few}, etc., have been introduced. With these techniques, model structure, parameters can be reused for building a model for similar or different problems. However, in all cases, retraining is needed. Also, there may be scenarios, as shown in Figure \ref{fig:overview}, where a faulty section of the network needs to be amputated or replaced. Now, suppose we can decompose a Convolutional Neural Network (CNN) into smaller components (modules), where each component can recognize a single output class. In that case, we can reuse these components in various settings to build a new model, remove a specific output class, or even replace an output class from a model without retraining. In the past, modular networks~\cite{andreas2016neural, hu2017learning, ghazi2019recursive}, capsule networks~\cite{hinton2000learning, sabour2017dynamic}, etc., have been studied to train the network and incorporate memory into the learning process. But, these modules are not created for enabling reusability and replaceability. 
As others have noted~\cite{pan2020decomposing}, 
there are strong parallels between deep neural network development now and software development
before the notion of decomposition was introduced~\cite{dijkstra1982role, parnas1972criteria, tarr1999n}, and developers wrote monolithic code that can be reused or replaced as easily. 

Recently, decomposition has been used to enable reuse and replacement in dense-based networks~\cite{pan2020decomposing}. 
This work shows that decomposed modules can be reused or replaced in various scenarios. However, this work focused on dense-based networks and did not explore a more complex set of deep learning techniques, e.g., convolutional neural networks. This work~\cite{pan2020decomposing} relies on the dense architecture of the models, where nodes and edges (weight and bias) have a one-to-one relationship. In contrast, edges in CNN are shared among all the input and the output nodes.

In this paper, we propose an approach to decompose a CNN model used for image classification into modules, in which each module can classify a single output class. Furthermore, these modules can be reused or replaced in different scenarios. In Figure \ref{fig:overview}, we illustrate an issue faced by users while using the Google Photo app. We show how Google has resolved the problem and how replacing and reusing decomposed modules can be applied. 
In the past, Google Photo App tagged a black woman as a gorilla~\cite{verge}.
To resolve this issue, Google decided to remove the output class ``Gorilla'' 
by suppressing the output label~\cite{verge}. Precisely, they have suppressed the output for ``Chimpanzee'', ``Monkey'', ``Chimp'', etc. Though the problem can be temporarily solved by suppressing the output label, to fix the model, one needs to retrain the model. We propose four different solutions based on reusing and replacing decomposed modules. We discuss each approach, its pros, cons, and illustrate how retraining the model can be avoided. The code and other artifacts are made available~\footnote{https://github.com/rangeetpan/Decomposition}.

The key contributions of our work are the following: 

\begin{itemize}
	\item We introduce a technique to decompose a CNN model into modules, one module per output class.  
	\item We describe an approach for reusing these decomposed modules in different scenarios to 
	build models for new problems.
	\item We replace a part of the model with decomposed modules.
	\item We evaluate our approach against the $CO_2e$ consumption of models created from scratch and reusing or replacing decomposed modules.
\end{itemize}

\textbf{Results-at-a-glance. }Our evaluation suggests that decomposing a CNN model can be done with a little cost (-1.77\% (top-1) and -0.85\% (top-5)) compared to the trained model accuracy. Reusing and replacing modules can be done when the modules belong to the same datasets (reusability: +2.5\%, replaceability: +0.7\% (top-1) and +1.0\% (top-5)) and the different datasets (reusability: -7.63\%, replaceability: +2.5\% (top-1) and -7.0\% (top-5)). Furthermore, enabling reusability and replaceability reduce $CO_2e$ emission by 37 times compared to training from scratch.

\textbf{Outline.} In \S\ref{sec:related} we describe the related work. Then in \S\ref{sec:methodology}, we discuss our approach to decompose a CNN model into modules. In \S\ref{sec:results}, we answer three research questions to evaluate the cost of decomposition, the benefit of reusing or replacing the modules, and the resource consumption. Lastly, in \S\ref{sec:conclusion}, we conclude.
\begin{figure*}[htp]
	\centering
	\includegraphics[, width=1\linewidth]{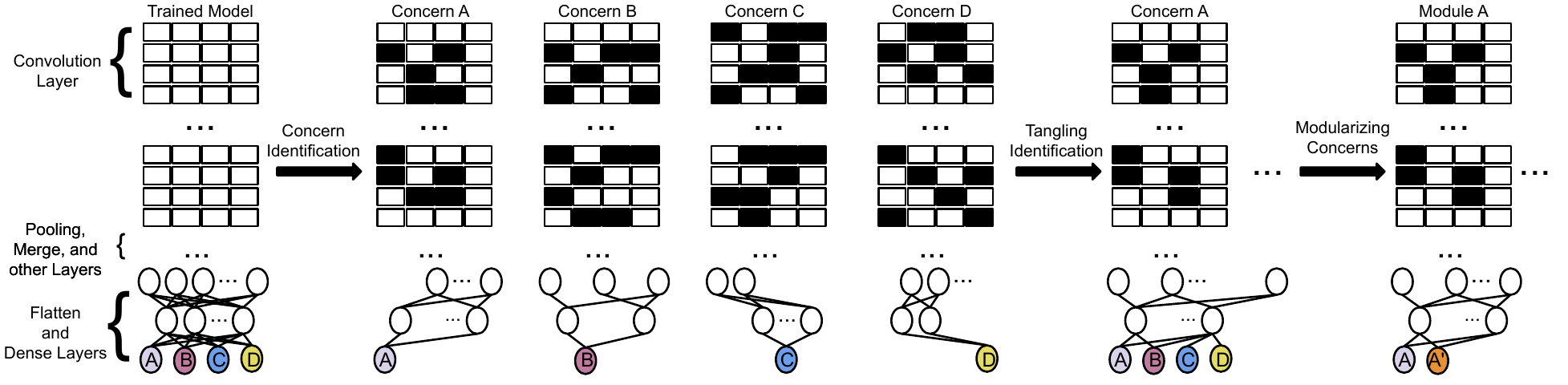}
	\caption{High-level Overview of Our Approach. Inactive nodes are denoted by black boxes.}
	\label{fig:approach}
\end{figure*}

%% file: tbMotiv.tex
\begin{tabular}{|p{1cm}|p{8.5cm}|p{2.8cm}|p{3.8cm}|}
	\hline
\multicolumn{1}{|c}{\textbf{Strategies}}                                                                      & \multicolumn{1}{|c|}{\textbf{Description}}                  & \multicolumn{1}{c}{\textbf{Pros}}          & \multicolumn{1}{|c|}{\textbf{Cons}}        \\
\hline
\hline
\multirow{1}{*}[-2.8em]{\textbf{Option 1}}                                                                        & We decompose the faulty CNN model into modules for each output class. Each such module is a binary classifier that classifies whether an input belongs to the output class (for which the module is decomposed) or not. Now, suppose we have another working model trained with the same or a subset of the dataset with the gorilla and person output labels. This new model does not exhibit the behavior present in the faulty one. Then, we can replace the faulty gorilla module with the working gorilla module.& Since the module belong to the same dataset as the faulty model, the module will have sufficient information to both recognize and distinguish new inputs. & These models are massive, and training is very sufficiently costly. Thus, the availability of a similar trained model with the Gorilla class may not be feasible for all conditions. \\
\hline
\multirow{1}{*}[-2.61em]{\textbf{Option 2}}                                                                    &                             We decompose the faulty model into modules. Moreover, we train a new model with a person and gorilla classes and validate that the trained model does not demonstrate faulty behavior. We decompose the newly trained model into two modules (gorilla and person) and replace the faulty gorilla module with the new decomposed gorilla module.  &     Training with a small dataset requires less resources in comparison to training the whole dataset.       & Decomposition can not alone solve the problem. However, the decomposition bundled with the traditional training could help in this situation.  \\
\hline
\multirow{1}{*}[-.61em]{\textbf{Option 3}}                                                                   &            We decompose the faulty model into modules and remove the gorilla module from the collection. In this scenario, there is no cost of retraining involved.              &    This method is cost-effective.  & The actual problem of faulty classification has not been addressed.\\
\hline        
\multirow{1}{*}[-1.61em]{\textbf{Option 4}}                                                                         &            We can also reuse the person and gorilla module from a working model without the faulty behavior. If any input is classified as a person or gorilla using the faulty model, we reuse the working modules to verify it further.           &   This approach is cost effective and used for further verification. & Since this approach involves another layer of verification using modules, there is an overhead present. \\
\hline                 
\end{tabular}

%% file: related.tex
\section{Related Works}
\label{sec:related}
There is a vast body of work~\cite{backus1957fortran, backus1960report, dijkstra2001go, dijkstra1970notes, dijkstra1982role, parnas1972criteria,  liskov1974programming, parnas1976design, america1990designing, dhara1997forcing, cardelli1997program, flatt1998units, tarr1999n} on software decomposition
that has greatly influenced us to decompose CNN model into reusable and replaceable modules.

The closest work is by Pan and Rajan~\cite{pan2020decomposing}, where the dense-based model has been decomposed into modules to enable reuse and replacement in various contexts. Though this work has motivated to decompose a CNN model into modules, the dense-based approach cannot be applied due to: 1) the shared weight and bias architecture in convolution layers, and 2) support for layers other than dense.
Also, this work did not evaluate the impact of decomposition on $CO_2e$ emission during training. 

Ghazi~\etal~\cite{ghazi2019recursive} have introduced modular neural networks to incorporate memory into the model. They have proposed a hierarchical modular architecture to learn an output class and classes within that output class. Though this work has been able to increase the interpretability of the network by understanding how inputs are classified, the modules are not built to enable reusability or replaceability. Other works on modular networks~\cite{andreas2016neural, hu2017learning} have learned how different modules communicate with others to achieve better training. Also, capsule networks~\cite{hinton2000learning, sabour2017dynamic} can be utilized to incorporate memory into deep neural networks. In capsule networks, each capsule contains a set of features, and they are dynamically called to form a hierarchical structure to learn and identify objects. However, modules decomposed by our approach can be reused or replaced in various scenarios without re-training.

\citeauthor{sairam2018hsd}~\cite{sairam2018hsd} have proposed an approach to convert a CNN model into a hierarchical representation. At each level, nodes representing similar output classes are clustered together, and each cluster is decomposed further down the tree. Finally, with parameter transfer, the tree-based model is trained. While, in that work, a sub-section of the tree-based can be reused for a subset of the output classes, our approach decomposes a trained CNN model into reusable and replaceable modules to build a model for a new problem (intra and inter dataset) without retraining. Furthermore, we have shown that reusing and replacing modules decreases CO2e consumption significantly.

%% file: methodology.tex
\section{Approach}
\label{sec:methodology}
In this section, we provide an overview of our approach for CNN model decomposition. We discuss the challenges in general. We also discuss each step of decomposing a CNN model into modules.

Figure \ref{fig:approach} shows the steps for decomposing a CNN model into modules. Our approach starts with the trained model and identifies the section in the CNN that is responsible for a single output class (Concern Identification). Since we remove nodes for all non-concerned output classes, the identified section acts as a single-class classifier. To add the notion of the negative output classes, we add a limited section of the inputs from unconcerned output classes (Tangling Identification). Finally, we channel the concerns to create a module(s) (Concern Modularization). In the example, we show the decomposition of the module for the output class A. Here,  a model trained to predict four output classes has been decomposed into four modules. Each one is a binary classifier that recognizes if an input belongs to the output class. 
In this paper, we use \textit{concerned} and \textit{unconcerned} as terminologies that represent the input belonging to the output class for which module has been created and all the other output classes, respectively.
For example, in Figure~\ref{fig:approach}, we show a module that is responsible for identifying output class $A$. For that module, output class $A$ is the concerned output class, and other output classes, e.g., $B$, $C$, and $D$, are the unconcerned classes.
\subsection{Challenges}
\label{subsec:challenges}
\begin{figure}
	\centering
	\includegraphics[, width=1\linewidth]{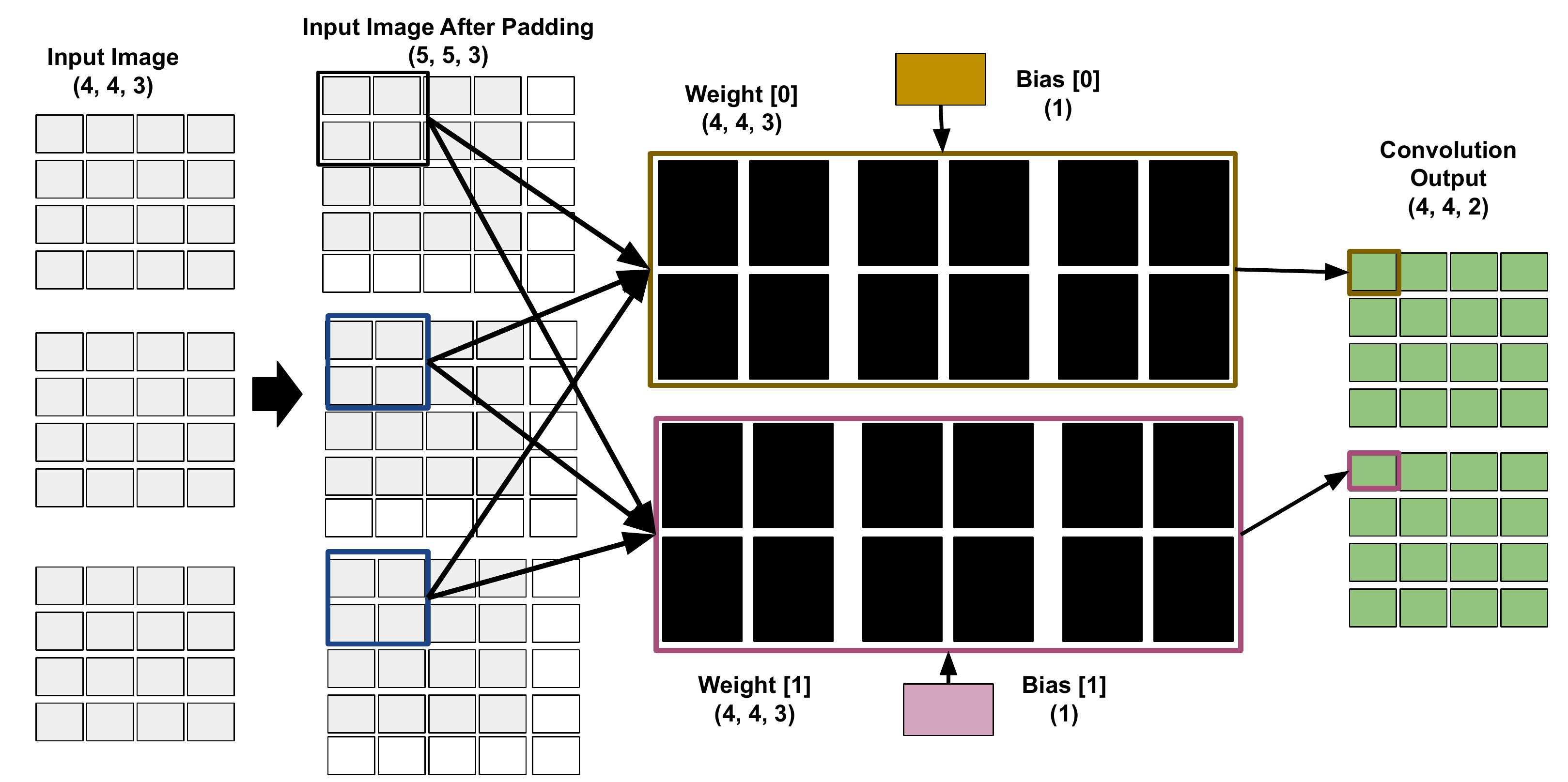}
	\caption{Architecture of a Convolutional Layer. }
	\label{fig:cnn}
\end{figure}

In the prior work on decomposing dense-based models, 
modules were created by removing edges. There, the value of the node has been computed. If it is $\le$0, then all incoming and outgoing edges are removed.
In a typical dense-based model, the first layer is a flatten layer that converts an input into a single-dimensional representation. From there, one or more dense layers are attached sequentially. In each such dense layer, nodes are connected with two different edges, 1) an edge that connects with other nodes from the previous layer (weight) and 2) a special incoming edge (bias). However, for a \conv layer, this is not the case. 
Figure \ref{fig:cnn} illustrates a traditional \conv layer. In that figure, on the left side, we have input nodes. Weight and bias are shown in the middle, and finally, on the right side, we have the output nodes. Each node in the input is not directly connected with the weight and bias, rather a set of nodes (sliding window) are chosen at a time as the input, and the output is computed based on that. The weight and bias are the same for all the input blocks. Due to these differences in the underlying architecture of a \conv layer, the dense-based decomposition approach could not be applied to a \conv layer directly. In the next paragraphs, we discuss each of such challenges.

\textbf{Challenge 1: Shared Weight and Bias.} 
The removal of edges in dense-based layers has been done to forcing the value of the nodes that are not needed to be 0 and eventually remove them from the decision-making process in the module. This approach is possible in dense layers because there is a one-to-one relationship between two nodes that belong to subsequent layers. If we remove a part of the weight and bias for one node in the  \conv layer, the weight and the bias will be turned off for all other nodes as well, and there will not be any output produced. 

\textbf{Challenge 2: Backtracking.} The prior work  channels the concerns to convert the module into a binary classification problem. However, before channeling the output nodes, a backtracking approach has been applied that starts with removing nodes at the last hidden layer that are only connected to the unconcerned nodes at the output layers and backtrack to the first hidden layer. However, the same approach cannot be directly applied to the CNN models due to the differences in the architecture of the convolutional layers and other supporting layers, e.g., \textit{Pooling}, \textit{Merge}.

\textbf{Challenge 3: Loss of Accuracy in Inter-Dataset Scenarios.} Prior work evaluated their approach by reusing and replacing modules that belong to different datasets. It has been found that such scenarios involve a non-trivial cost.
Since these modules involved in the reuse and replace scenario do not belong to the same datasets, they are not programmed to distinguish between them. To remediate the cost, we propose a continuous learning-based approach that enables retraining of the decomposed modules.

%% file: technique1.tex
\subsection{Concern Identification}
Concern Identification (CI) involves identifying the section of the CNN model, responsible for the single output class. As a result of this process, we remove nodes and edges in the model. In traditional CNN models for image classification, both convolution, and dense layers have the notion of node and edges, and we discuss the concern identification approaches for both the layers. 

\noindent\textbf{Dense Layers:}
Here, the concerned section can be identified by updating or removing the edges connected to the nodes. In a dense-based network, there is a one-to-one relationship between the edges connecting the nodes from different layers (including the bias nodes). For each edge, the originating and the incident nodes are unique except for bias, where the incident node is unique. The edges between nodes are removed or updated based on the value associated with the nodes. In this process, we identify nodes based on the assumption that ReLU has been used as the activation function. Since our work is focused on image-based classification models, ReLU is the most commonly used activation function for the hidden layers. Also, prior work on decomposition~\cite{pan2020decomposing} has been carried out with the same assumption.

First, we compute the value associated with the nodes by applying training inputs from the concerned output class. For an input, if the computed value at a node is $\le$0, then we remove all the incident and originated edges. We do that for all the concerned inputs from the training dataset. If the value associated with a node is $>$0 for some input and $\le$0 for other inputs, we do not remove that node. 
For instance, for a layer $L_d$, there are $n_{L_d}$ number of nodes, and the preceding and the following layer has $n_{L_{d-1}}$ and $n_{L_{d+1}}$ nodes, respectively. For any node at the layer $L_d$, there will be $n_{L_{d-1}}$incident edges and $n_{L_{d+1}}$ outgoing edges. Based on our computation, if a node $n_i$ is inactive (value $\le$0), then all the incoming and outgoing weight edges ($n_{i(L_{d-1})}$ + $n_{i(L_{d+1})}$) and one bias edge incident to $n_i$ will be removed. We do the same for all the hidden dense layers. 
\input{algoci.tex}

\noindent\textbf{Convolution Layers:}
In a convolutional layer, we identify the inactive sections in the output nodes by using a mapping-based technique that stores the position of the nodes that are not part of a module. 
In Algo.\ref{algo:concern}, we describe the steps involved in building
the map and storing the nodes' positions. First, we store the
weight and bias for all the layers. Then, we identify the parts of the
convolution layer that are not used for a single output class. 
We start by computing all possible combinations of sliding windows at line \ref{algo1:10}. To build the sliding windows, we use the $stride$, $padding$ as input. Below, we describe each such parameter.

\textbf{Sliding Window.} In convolutional layers, instead of one input node at a time, a set of nodes are taken as an input for computation. For instance, in Figure \ref{fig:cnn}, the blue box is a sliding window. For each sliding window, one or more output nodes are created based on the size of the shared weight in the layer.

\textbf{Padding.} Two variations of padding is possible in CNN, zero-padding and with-padding. In zero-padding, the input is not changed. For with-padding, the input is padded based on the size of the sliding window, and the size of the output will be the same as the input. For the example shown in Figure \ref{fig:cnn}, we used the with-padding, and that adds padding with value zero and transforms the input into (5, 5, 3) size (the white boxes are the added padding).

\textbf{Stride.} This parameter controls the movement of the sliding window while computing the output. Stride along with the padding decides the output of the layer.

Once we compute the sliding windows (line \ref{algo1:10}), we feed inputs from the training dataset to our approach and observe the output value of that particular convolution layer. At line \ref{algo1:11}, we compute the output of the convolution layer based on S*W + B, where S, W, and B denote the sliding window, shared weight, and bias, respectively. Then, we monitor the value at each node (line \ref{algo1:12}-\ref{algo1:25}). If a node has a value $\le0$, we store the position of the node in our map. We initialize the map with all the nodes (for the first input) that have value $\le0$ (line \ref{algo1:14}-\ref{algo1:16}). The \textit{first} flag is used to denote this first input to the module. Then, we remove the nodes that are previously in the mapping list, but the nodes have a positive value for the input under observation (line \ref{algo1:17}-\ref{algo1:24}). We perform such operations to identify the section of the layer that is inactive for a particular concern.
For the batch normalization layer, there is no weight or bias involved, and the layer is utilized for normalizing the input based on the values learned during the training session. Max pooling and average pooling are utilized for reducing the size of the network using the pool size. For merge or add layer, we add the value computed from the two layers connected with this layer.

%% file: algoci.tex
\begin{algorithm}[ht]
	\caption{Concern Identification (CI).}
	\footnotesize
	\label{algo:concern}
	\begin{algorithmic}[1]
		\Procedure{Initialization($model$)}{}\label{algo1:1}
		\State $convW$, $convB$ =[]\label{algo1:2}
		\For{each layer $\in model$}\label{algo1:3}\Comment{Retrieve the weight and bias}
		\If{$layer_{type}$=="Convolution"}\label{algo1:4}
		\State $convW.add(layer.Weight); convB.add(layer.Bias);$
		\Else \If{$layer_{type}$=="Dense"}\label{algo1:5}
		\State  $denseW.add(layer.Weight); denseB.add(layer.Bias);$\label{algo1:6}
		\EndIf
		\EndIf
		\EndFor
		\State \Return $convW, convB, denseB, denseW$\label{algo1:7}
		\EndProcedure
		\Procedure{CILayer ($model$, $input$, $convW_{Layer}$,\label{algo1:8} $convB_{Layer}$,  $convMap_{Layer}$, $pad$=$with$, $Stride$=$1$, $first$=$False$)}{}
		\State $I$=$input$\label{algo1:9}
		\State $sliding_{w}$=$procSliding(I$, $pad$, $(Stride$, $Stride))$\label{algo1:10} \Comment{Sliding window}
		\State $Output$=$sliding_{w}*convW_{Layer}$ + $convB_{Layer}$\label{algo1:11} 
		\State $flatOutput$=$flatten(Output)$\label{algo1:12} \Comment{Convert into an 1-D array}
		\For{$j=0$ to $j=|flatOutput|$}\label{algo1:13} 
		\If{$first$}\label{algo1:14}
		\If{$flatOutput[j]$<=0}\label{algo1:15}\Comment{Identify the inactive nodes}
		\State $convMap_{Layer}.add(j)$\label{algo1:16}
		\EndIf
		\Else \Comment{Remove the inactive node if it is active for other inputs}
		\If{$j\in convMap_{Layer}$ }\label{algo1:17}
		\If{$flatOutput[j]$>0}\label{algo1:18}
		\State $index$=$findIndex(flatOutput, j)$\label{algo1:19}
		\State temp=[]\label{algo1:20}
		\For{$k=0$ to $k=convMap_{Layer}$}\label{algo1:21}
		\If{$k!=index$}\label{algo1:22}
		\State $temp.add(convMap_{Layer}[k])$\label{algo1:23}
		\EndIf
		\EndFor
		\State $convMap_{Layer}$=$temp$\label{algo1:24}
		\EndIf
		\EndIf
		\EndIf
		\EndFor
		\State \Return $convMap_{Layer}$, $Output$\label{algo1:25}
		\EndProcedure
		\Procedure{CI ($model$, $input, convMap$)}{}
		
		\State $convW, convB, denseB, denseW$=$initialization(model)$\label{algo1:26}
		\For{each $layer \in model$}\label{algo1:27}\Comment{Perform CI for all the layers}
		\State count=0\label{algo1:28}
		\If{$layer_{type}$==$``Convolution"$}\label{algo1:29}
		\State convMap[count], output = CILayer(($model$, $input$, $convW[count]$, $convB[count]$,  $convMap[count]$, $pad$=$layer_{pad}$, $Stride$=$layer_{stride}$))\label{algo1:30}
		\EndIf
		\If{$layer_{type}$=$``BatchNormalization"$}\label{algo1:31}
		\State output = BatchNorm(($model$, $Gamma$))\label{algo1:32}
		\EndIf
		\If{$layer_{type}$==$``AveragePooling"$}\label{algo1:33}
		\State output = AvgPool(($model$, $Pool_{Size}$))\label{algo1:34}
		\EndIf
		\If{$layer_{type}$=$``MaxPooling"$}\label{algo1:35}
		\State output = MaxPool(($model$, $Pool_{Size}$))\label{algo1:36}
		\EndIf
		\If{$layer_{type}$=$``Add"$}\label{algo1:37}
		\State output = input+$Previous_{input}$\label{algo1:38} \Comment{Add both the layers that are merged}
		\EndIf
		\If{$layer_{type}$=$``Dense"$}\label{algo1:39}
		\State output = denseMod($model$, $input$, $indicator$=$False$, $denseW[count]$, $denseB[count]$)\label{algo1:40}\Comment{Apply dense-based CI}
		\EndIf
		\If{$layer_{type}$=$``Flatten"$}\label{algo1:41}
		\State output = flatten($input$)\label{algo1:42} \Comment{Apply flatten-based CI}
		\EndIf
		\State $input$=$output$
		\EndFor
		\EndProcedure
	\end{algorithmic}    
\end{algorithm}


%% file: technique3.tex
\subsection{Tangling Identification}
In concern identification, a module is created based on identifying the nodes and edges for a single output class. Since all the nodes and edges related to the other output classes in the dataset have been removed, the module essentially characterizes any input as the concerned output class or behaves as a single-class classifier. To add the notion of unconcerned output classes and able to distinguish between the concerned and unconcerned output classes, we bring back some of the nodes and edges to the module. In Tangling Identification (TI), a limited set of inputs belonging to the unconcerned output classes have been added. Based on the prior work, we add concerned and unconcerned inputs with a 1:1 ratio. For instance, if we have a problem with 200 output classes and build a module based on observing 1000 inputs from the concerned class, then we observe 5 inputs from each unconcerned class (5x199=995$\approx$1000).
\subsection{Modularizing Concerns}
\label{subsec:cm}

So far, we identify the section of the network that is responsible for an output class and added examples from unconcerned output classes. However, the module is still an $n$-class classification problem. We channel the output layer for each module to convert that into a binary classification-based problem. 
 However, before applying the channeling technique, we remove irrelevant nodes (do not participate in the classification task for a module) using a bottom-up backtracking approach. 

\noindent\textbf{Dense Layers:}
We channel the out edges as described by the prior work~\cite{pan2020decomposing}. Instead of having $n$ nodes at the output layer ($L_{d}$, where $d$ is the total number of dense-based layers), two nodes (concerned and unconcerned output nodes) have been kept. For instance, the concerned output class for a module is the first output class in the dataset. We have $n$ output classes and $n$ nodes ($V_1, V_2, \dots, V_n$) at the output layer. Also, the layer preceding the output layer ($L_{d-1}$) has $n_{L_{d-1}}$ nodes. For each node at the output layer, there will be $n_{L_{d-1}}$ incident edges. For instance, the incoming edges for $V_1$ node will be $E_{11}, E_{21}, \dots, E_{n_{L_{d-1}}1}$, where $E_{n_{L_{d-1}}*1}$ (in this case, n=1) denotes that an edge is connected with $n_{L_{d-1}}^{th}$ node from $L_{d-1}$ layer and the first node at the output layer. For the module responsible for identifying the first output label, the edges incident to the first node (as the concerned node is $V_1$) at the output layer have been kept intact. However, all the other edges are modified. All the edges incident to any of the unconcerned nodes ($V_2$, $V_3$, $\dots$, $V_n$) at the $L_d$ layer will be updated by a single edge. The assigned weight for the updated edge is the mean of all the edges (same for bias). Then, that updated edge has been connected to a node at the output layer, which is the unconcerned node for the module. For a module, there will be two nodes at the output layer, $V_c$ and $V_{uc}$, where $V_c$ and $V_{uc}$ denote the concerned node and the unconcerned node. All the updated edges will be connected to the $V_{uc}$. 

\noindent\textbf{Modularizing Concern: Backtrack:}
Once the nodes at the output layer are channeled, we backtrack the concerned and unconcerned nodes to the previous layers. In this process, we remove the nodes that only participate in identifying the unconcerned classes. First, we discuss the backtracking approach to handle the dense layers and other non-dense layers, and then we describe how we can backtrack till the input to reduce the size of the module. In all approaches, we support the other layers, e.g., pooling, merge, flatten.

\begin{figure}[htp]
	\centering
	\includegraphics[, width=0.88\linewidth]{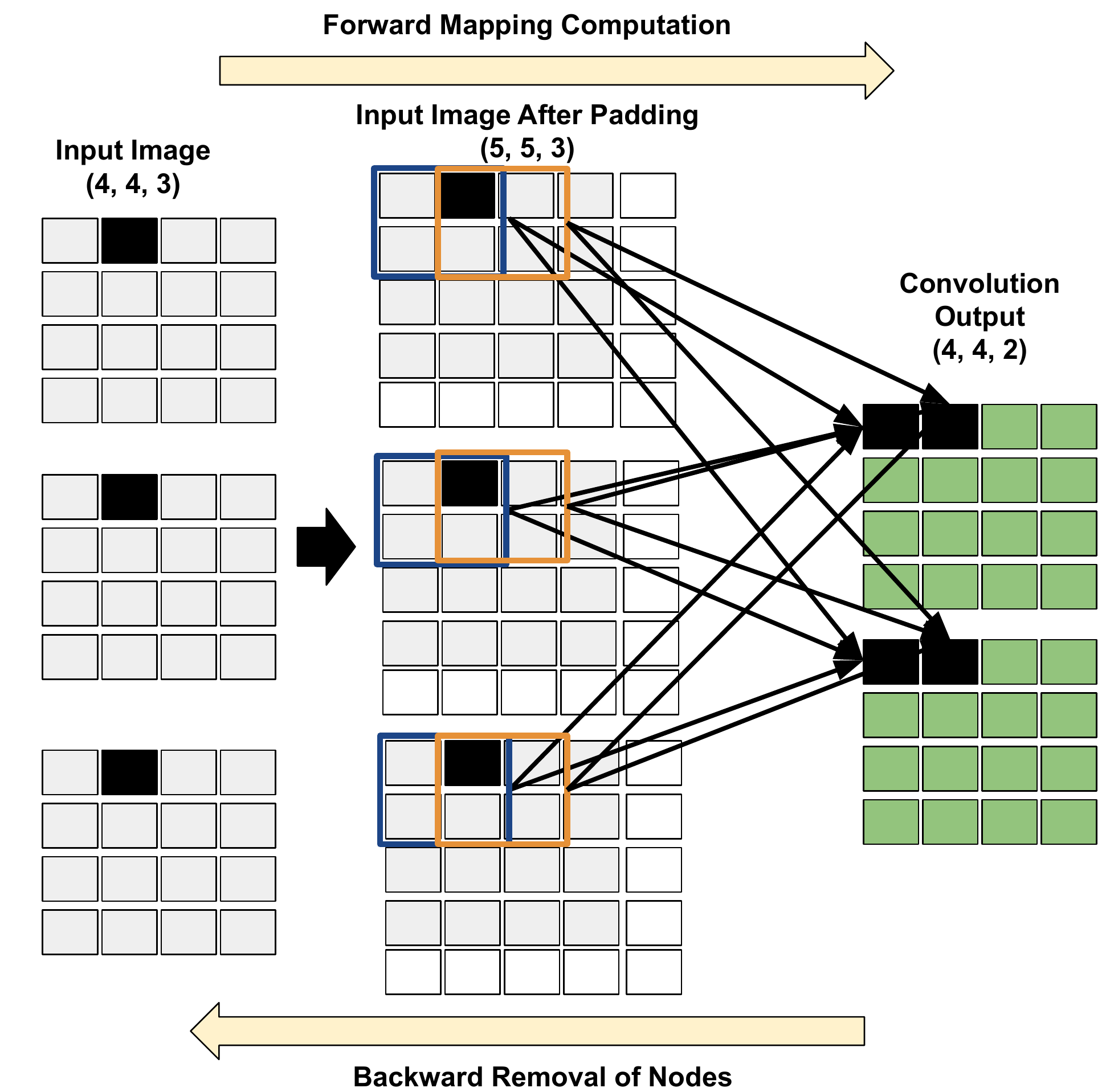}
	\caption{Backtrack Through a Convolution Layer. }
	\label{fig:backtrack}
\end{figure}
\input{algodbp.tex}

\textbf{Modularizing Concern: Backtrack To Last Convolutional Layer (MC-BLC).}
Once we channel the output layer, we prune the network based on the inactive nodes at the dense layers. 
In a typical CNN model, either a \textit{Pooling} layer, \textit{Convolution} layer, or a \textit{Merge} layer will precede \textit{Dense} layers and the \textit{Flatten} layers 
In this approach, we leverage that information to backtrack through the \textit{Dense} layers (including \textit{Flatten} layer) to the last convolution layer(s) (based on the presence of \textit{Merge} layer or not) or the pooling layer. In this process, we identify the nodes that have edges $E=\left\{E_{ij}; i\in L_{d-1}, j\in L_{d}, j\in V_{l(uc)} \right\}$, where all the edges are only connected to the unconcerned nodes ($V_{uc}$) at the $d^{th}$ layer (line \ref{algo3:4}-\ref{algo3:10}). We start the backtracking process from the output layer and move through the dense layer. For each layer, we identify the nodes connected to the unconcerned nodes in the following layer and remove them from the model. Also, we tag the removed nodes as the unconcerned nodes for that layer. To identify the nodes that strongly contribute to the unconcerned node, we introduce a constant ($\delta$) to verify the value associated with the node. Based on the experimental evaluation, we used $\delta=0.5$. Then, we backtrack the nodes at the layer preceding the output layer is identified at line \ref{algo3:11}-\ref{algo3:17}. Finally, we backtrack to the flatten layer. In a traditional CNN model for image classification, the flatten layer is preceded by a pooling layer, or a merge layer, or a convolutional layer. 
If the preceding layer is a convolution layer, we update the mapping (as discussed in \S\ref{subsec:cl}) for that particular convolutional layer. Since the convolution layer's output is directly reshaped into the flatten layer, there is a one-to-one relationship between the two layers. If the preceding layer is a pooling layer, then there will be $X^2$ (X is the pool size) inactive nodes at the pooling layer for one inactive node at the flatten layer. If the preceding layer is a merge layer, then the convolution layers that are merged will be updated. 

\textbf{Modularizing Concern: Backtrack To Input (MC-BI).}
In the previous approach, we can only backtrack from the output layer to the last convolution layer. However, we cannot backtrack through the convolution layer. In a convolution layer, the input nodes cannot be directly mapped with the output nodes. For instance, in Figure \ref{fig:backtrack}, the input image shown on the left side is turned into the image shown in the middle, which is after adding the paddings (for this example, we choose \textit{Valid} padding). In the output, the nodes on the top left corner for both arrays will be produced by the first sliding window (blue box) from each array shown in the middle. So, for mapping, a node in the output on the right side of the image, at least 4 (in this example, we chose the sliding window size to be 2x2) nodes can be mapped. Those four individual nodes are also mapped with other nodes in the output. The black-colored node in the middle is a part of two sliding windows (the blue box and the orange box). To remove irrelevant nodes from the convolution layer, we take a two-pass approach. First, we store the position of the nodes in each sliding window with the nodes in the output (forward pass). 
During the forward pass, we store the mapping $M=\left\{(V_i, V_j) ; V_j=f(V_i, W, B)\right\}$, where $f$ denotes the convolution operation. During the backward pass, we remove the nodes.
\input{algobp.tex}

In Algo. \ref{algo:cmfb}, we describe the step to do the mapping. From line \ref{algo4:1}-\ref{algo4:10}, the forward pass has been described. In the forward pass, we store the mapping between the sliding window and output nodes. In order to denote the position of the sliding window, we mark each node with a unique number (line \ref{algo4:4}-\ref{algo4:5}) before adding the padding. For padding, the nodes are marked as ``0'' as they are not present in the input of the \conv layer. Then for each output node, the input nodes are stored in a list. In this process, we define an operation named \texttt{sw} that computes the sliding windows from the weight $W$, padding $pad$, and stride. 
Now, we compute the mapping with the input and the output nodes at line \ref{algo4:12}. Then, we separate the input and the output nodes at line \ref{algo4:13} and \ref{algo4:14}. We scan through the inactivate nodes in the output and identify if they match the pattern as illustrated in Figure~\ref{fig:backtrack}. We identify all the input nodes that are mapped with an output node and vice versa. We focus on searching nodes that are not part of the padding operation and remove the nodes marked with 0 at line \ref{algo4:18}. For each such input node, we find all the output nodes generated from the particular input node. If these output nodes are already in the deactivation node list, we add the input node to the deactivation list for the preceding convolution (output of the preceding convolution layer is the input of the next convolution layer). If the previous layer is an add layer, then the two convolution layers that are merged at the merge layer are updated with the changes. If the preceding layer is a pooling layer, the update is carried based on the pool size.

\subsection{Continuous Learning}
\label{subsec:cl}
\begin{figure}
	\centering
	\includegraphics[, width=0.84\linewidth]{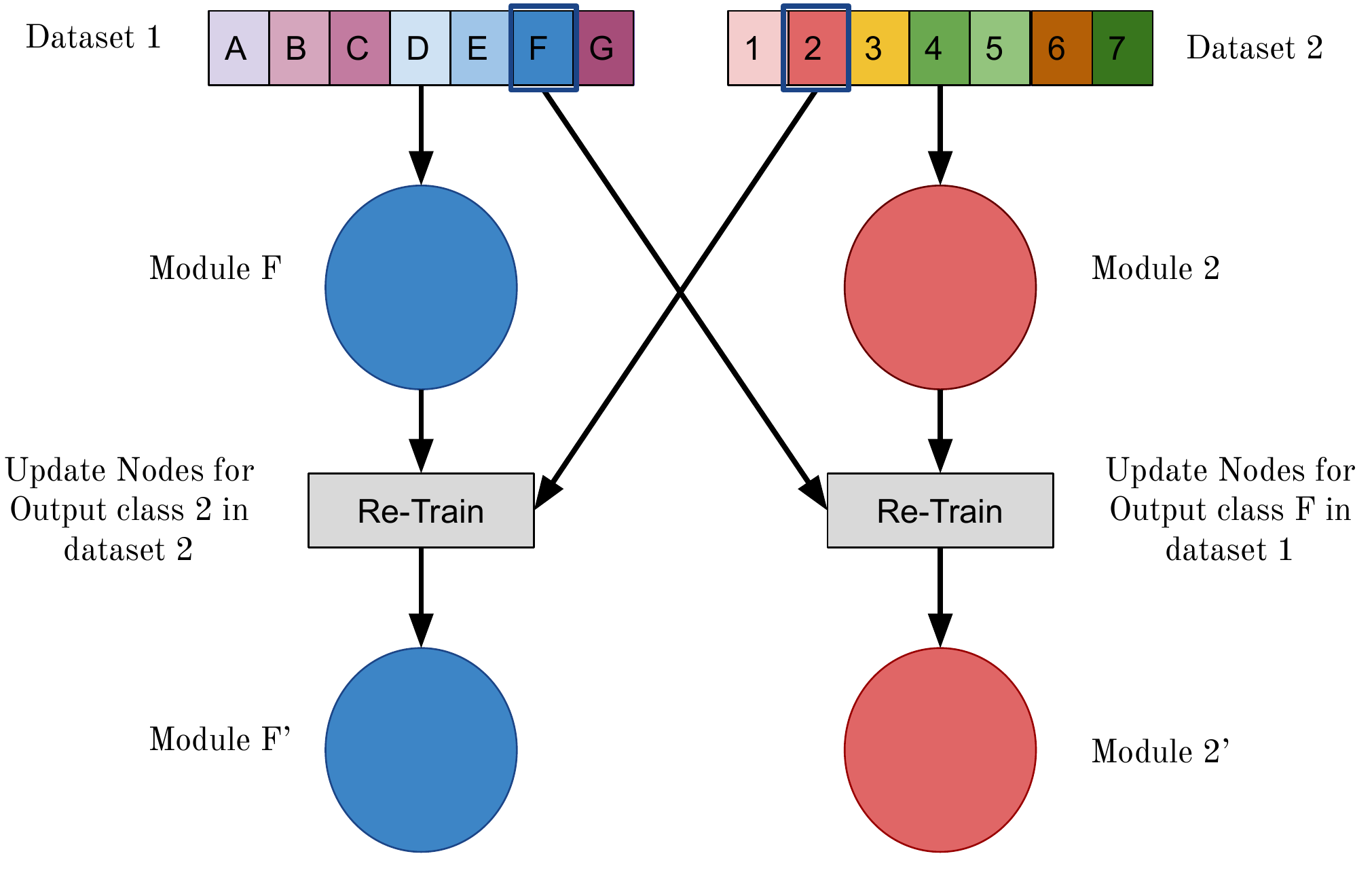}
	\caption{Continuos Learning. }
	\label{fig:cl}
\end{figure}

In the prior study, reusing modules that originated from different datasets involves non-trivial cost. Our intuition is that since the modules are originated from a different dataset, they still have some traits of the parent dataset. In fact, by applying the tangling identification approach, we deliberately add some unconcerned examples to learn the modules on how to distinguish between the concerned and the unconcerned examples. However, in the inter dataset scenarios, the unconcerned output classes are not the same. To solve this problem, we propose a continuous learning-based approach. In deep learning, continuous learning~\citeauthor{collobert2008unified}~\cite{collobert2008unified} has been widely applied. In Figure \ref{fig:cl}, we illustrate a reuse scenario, where module F is originated from dataset 1 and module 2 is originated from dataset 2. Dataset 1 represents a set of English letters (A-G), and applying decomposition creates modules for each output class. Similarly, dataset 2 represents a set of English digits (1-7), and decomposition creates seven modules, each for one output class. When module F and module 2 are reused in a scenario, based on the prior work, each input belongs to 2, and F will be given as input to the composition of the decomposed modules. However, due to the parent dataset traits, module 2 can recognize itself but does not know how to distinguish from any input belonging to the output class F. To learn the concerned output classes in this scenario, we take the unconcerned section of the dataset. For instance, for module F, the unconcerned output class will be output class 2 from dataset 2. We take the examples from output class 2 from dataset 2 and update module F by removing the nodes responsible for detecting the output class 2. We do the same for module 2, where we remove the nodes responsible for recognizing output class F from dataset 1. Finally, the modified modules are ready to be reused.

%% file: algodbp.tex
\begin{algorithm}[ht]
	\caption{MC-BLN: Modularizing Concern: Backtrack to Last Convolutional Layer (MC-BLN).}
	\footnotesize
	\label{algo:cmdbb}
	\begin{algorithmic}[1]
		\Procedure{CMBLN($model$, $num\_of\_labels$, $module\_class$, $UpdatedW$, $UpdatedB$)}{}\label{algo3:1}
		\State $temp$, $tempL$, $tempPool=[]$\label{algo3:2} \Comment{UpdateW: Weight for Dense Layers}
		\State $DenseLength=|UpdatedW|$\label{algo3:3}\Comment\Comment{UpdateB: Bias for Dense Layers}
		\For{$i=0$ to $i=UpdatedW[DenseLength-1][0]$}\label{algo3:4}
		\If{UpdatedW[DenseLength-1][i,$v_{uc}$]<=-$\delta$ and UpdatedW[DenseLength-1][i,$v_c$]>=+$\delta$}\label{algo3:6} 
		\State $temp.add(i)$\label{algo3:7} \Comment{$v_c$: concerned nodes, $v_{uc}$: unconcerned nodes}
		\EndIf
		\EndFor
		\State $tempL.add(temp)$\label{algo3:10}
		\For{$i=DenseLength-1|$ to $i=0$}\label{algo3:11}\Comment{Backtrack till the first Dense}
		\State $temp=[]$\label{algo3:12}
		\For{$j=0$ to $j=UpdatedW[i-1][0]$}\label{algo3:13}
		\If{$j \in tempL[i]$}\label{algo3:14}
		\If{UpdatedW[i-1][j:,]<=-$\delta$}\label{algo3:15}
		\State $temp.add(i)$\label{algo3:16}\Comment{Add nodes that are only connected with $v_{nc}$}
		\EndIf
		\EndIf
		\EndFor
		\State $tempL.add(temp)$\label{algo3:17}
		\EndFor
		\For{$i \in tempL[|tempL-1|]$}\label{algo3:18}
		\For{$j=0$ to $j=i*poolsize^2+j$}\label{algo3:19}\Comment{Convert to pooling layer}
		\If{Layer-1=="Convolution"}\label{algo3:20}
		\If{j not $\in$ convMap[depth]}\label{algo3:21}
		\State convMap[depth].add(j)\label{algo3:22} \Comment{Update convolution layer map}
		\EndIf
		\EndIf
		\If{Layer-1=="Add"}\label{algo3:23} \Comment{Update two convolution layer maps}
		\If{j not $\in$ convMap[depth]}\label{algo3:24}
		\State convMap[depth].add(j)\label{algo3:25}
		\EndIf
		\State depthAdd2=Add.Input2\label{algo3:26}
		\If{j not $\in$ convMap[depthAdd2]}\label{algo3:27}
		\State convMap[depthAdd2].add(j)\label{algo3:28}
		\EndIf
		\EndIf
		\EndFor
	
		\EndFor
		\EndProcedure
	\end{algorithmic}    
\end{algorithm}

%% file: algobp.tex
\begin{algorithm}[ht]
	\caption{MC-BI: Modularizing Concern: Backtrack to Input.}
	\footnotesize
	\begin{algorithmic}[1]
		\Procedure{Sliding\_Window\_Mapping($input$, $W$, $pad$, $stride$)}{}
		\State mapping =[], count=0\label{algo4:1}\Comment{Performs the forward pass and map input output nodes}
		\State $temp=zeros_like(input)$\label{algo4:2}
		\State $temp=temp.flatten()$\label{algo4:3}
		\For{$i=0$ to $i=|temp|$}\label{algo4:4}
		\State $temp[i]=i+1$\label{algo4:5}
		\EndFor
		\State $temp=temp.flatten()$\label{algo4:6}
		\State $sliding\_window=sw(temp, W, pad, stride)$\label{algo4:7}
		\For{$i=0$  to $i=|sliding\_window|$}\label{algo4:8}
		\For{$j=0$  to $j=W.length[3]$}\label{algo4:9}
		\State $mapping.add(temp[i][j], count)$\label{algo4:10}
		\EndFor
		\EndFor
		\EndProcedure
		\Procedure{CMBI($input$, $W$, $pad$, $stride$, $B$, $preceeding\_layer$, $deactive\_map$)}{}\label{algo4:11}
		\State $convDepth=depth$\label{algo4:12}
		\State $mapping\_window=Sliding\_Window\_Mapping$($input$, $W$, $pad$, $stride$, $B$)
		\State $source\_mapping=mapping\_window[:0]$\label{algo4:13}\Comment{Input nodes}
		\State $sink\_mapping=mapping\_window[:1]$\label{algo4:14}\Comment{Output nodes}
		\For{each $deactived\_node$ $\in deactive\_map$}\label{algo4:15}
		\State $source$=$source\_mapping$[$sink\_mapping$==$deactived\_node]$\label{algo4:16}
		\State $flag= True$\label{algo4:17}\Comment{Identify the source, where sink is deactive}
		\State $source\_mapping=source\_mapping[source\_mapping]>0]$\label{algo4:18}
		\For {$source\_node$ $\in source\_mapping$}\label{algo4:19}
		\If{$flag==true$}:\label{algo4:20}
		\State $sink\_node$=$sink\_mapping$[$source\_mapping$==$source\_node]$\label{algo4:21}
		\If{$|sink\_node-deactive\_map|>0$}\label{algo4:22}
		\State $flag=false$\label{algo4:23}\Comment{All sinks formed by source are not deactive}
		\EndIf
		\EndIf
		\If{$flag==true$}\label{algo4:24}
		\For{each $source$ $\in source\_mapping$}\label{algo4:25}
		\If{$source$ not $\in deactive\_map$}\label{algo4:26}
		\State $deactive\_map[-1].add(source+1)$\label{algo4:27}\Comment{Update Map}
		\EndIf
		\If{$preceeding\_layer$=="Add"}\label{algo4:28}
		\State $deactive\_map[-2].add(source+1)$\label{algo4:29}\Comment{Update Map}
		\EndIf
		\EndFor
		\EndIf
		\EndFor
		\EndFor
		\EndProcedure
	\end{algorithmic}   
\label{algo:cmfb} 
\end{algorithm}

%% file: results.tex
\section{Evaluation}
\label{sec:results}
\input{tbrq1.tex}
In this section, we discuss the experimental settings. Furthermore, we discuss the performance of decomposing the convolutional neural network into modules by answering three research questions, 1) does decomposition involve cost?, 2) how module reuse and replacement can be done compared to retraining a model from scratch?, and 3) does reusing and replacing the decomposed modules emit less $CO_2$?
\subsection{Experimental Settings} 
\subsubsection{Datasets} We evaluate our proposed approach based on three widely used and well-vetted datasets.

\noindent\textbf{CIFAR-10 (C10)~\cite{krizhevsky2009learning}:} This dataset comprises of 3-D images of different objects. It has 10 output classes, and the dataset is divided into training and testing datasets. The training dataset has 50,000 images, and the testing dataset has 10,000 images.  

\noindent\textbf{CIFAR-100 (C100)~\cite{krizhevsky2009learning}:} This dataset has 100 output classes. This is widely used to evaluate CNN-based studies due to the complexity of the dataset. Here, there are 50,000 training images and 10,000 testing images. The image size is similar to the CIFAR-10.

\noindent\textbf{ImageNet-200 (I200)~\cite{le2015tiny}:} This dataset is very popular and widely used to measure the scalability of CNN-based applications. Unlike the previous two datasets, the total number of output classes is 200. 
ImageNet dataset has been designed to help the computer-vision related task in machine learning. 
This dataset has 80,000+ nouns (name of the output class), and for each class, there are at least 500 images associated with it. This dataset has been used in training models in real-life scenarios. However, due to the complexity of the dataset, a smaller dataset with similar complexity has been made public for research purposes. This dataset (ImageNet-tiny) comprises 200 types of images. The training dataset has 100000 images, and the testing has 10000 images. 
\subsubsection{Models} We used the ResNet~\cite{he2016deep, he2016identity} models to evaluate our proposed approach. In a ResNet model, there are multiple blocks present and each block consists of \texttt{Convolution}, \texttt{Add}, and \texttt{Activation} layer. There are two versions of ResNet, the original version, where there is a stack of \texttt{Convolution}, \texttt{Add}, and \texttt{Batch Normalization} layers are put together to form a residual block, and those residual blocks build the network. In the second version, the residual block is modified to form a bottleneck layer with \texttt{Convolution}, \texttt{Add}, and \texttt{Batch Normalization}. We performed our evaluation against the first version of the network for simplicity and to reduce the training time and resources. Each such version can either have \texttt{Batch Normalization} layer or not as described in the original paper. Since \texttt{Batch Normalization} is only used to train and does not participate in the prediction, we chose the model without \texttt{Batch Normization} to reduce the training time as some of our experiments include training models multiple times. We used ResNet-20, ResNet-32, and ResNet-56, where 3, 5, and 9 residual blocks are present with 21, 33, and 57 convolution layers, respectively. The reported model accuracies are different from the original paper, as we trained the model from scratch with 200 epochs.
\subsubsection{Metrics Used}

\textbf{Accuracy.} We compute the composed accuracy of the decomposed modules as shown in the prior work~\cite{pan2020decomposing}. Pan~\etal computed the top-1 accuracy for all the experiments, whereas we compute both top-1 and top-5 accuracies. 

\textbf{Jaccard Index.} Similar to the prior work, we compute the variability between the model and the modules using Jaccard Index.


\textbf{$\boldsymbol{CO_2e}$ Emission.} In this context, we refer to $CO_2e$ as carbon dioxide and equivalent gases. These gases are harmful to humankind and the earth due to various reasons. One of them is that they are heat-trapping gases, and the excess presence of such gases can increase the average temperature of the earth, which can lead to several devastating circumstances. To measure the total emission of such gases due to computation, we utilize the metrics used by Strubell~\etal~\cite{strubell2019energy}. The total power consumption during the training is measured as ,
\begin{equation}
\label{eq:1}
p_t=\frac{1.58t(p_c+p_r+gp_g)}{1000}
\end{equation}
In this equation, $p_c$, $p_r$, $p_g$, and $g$ denote the average power consumption of CPU, DRAM, GPU, and the total number of GPU cores, respectively. The $t$ denotes the time. Here, 1.58 denotes the PUE co-efficient, which is the same value used in the prior work. We performed our experiment on iMac with 4.2 GHz Quad-Core Intel Core i7 and 32 GB 2400 MHz DDR4 RAM. Since we do not have any GPU, both $g$ and $p_g$ are zero. The power consumption has been measured using Intel Power Gadget
~\cite{intel}. 
Finally, the $CO_2e$ emission has been computed as,
\begin{align}
CO_2e=0.954p_t
\end{align}

%% file: tbrq1.tex
\begin{table*}[htp]
	\centering
	\caption{Decomposition Effectiveness and the Variability between the Decomposed Modules and the Trained Model.}
\footnotesize
\label{tb:rq1}
\begin{tabular}{|l|c|c|c|c|c|c|c|c|c|c|c|}
\hline
\multicolumn{1}{|c|}{}                        & \multicolumn{2}{c|}{\textbf{Acc}} & \multicolumn{3}{c|}{\textbf{CI+TI+MC}} & \multicolumn{3}{c|}{\textbf{CI+TI+MC-BLN}} & \multicolumn{3}{c|}{\textbf{CI+TI+MC-BLN+MC-BI}} 
\\ 
\cline{2-12} 
\multicolumn{1}{|c|}{\multirow{-2}{*}{\textbf{Model}}} & \textbf{Top 1}              & \textbf{Top 5}              & \textbf{Top 1}      & \textbf{Top 5}      & \textbf{JI}     & \textbf{Top 1}        & \textbf{Top 5}        & \textbf{JI}        & \textbf{Top 1}        & \textbf{Top 5}        & \textbf{JI}        \\ \hline
\hline
\textbf{CIFAR10-R20}                                                  & 87.1\%                        & 99.5\%                           & 86.9\%                & 96.0\%                   & 0.75            & 87.0\%                  & 99.4\%                     & 0.75               & 86.8\%                  & 95.8\%                     & 0.75               \\ \hline
\textbf{CIFAR10-R32}                                                  & 85.3\%                        & 99.2\%                           & 78.4\%              & 96.0\%                   & 0.51            & 78.6\%                  & 98.1\%                     & 0.51               & 78.4\%                  & 97.0\%                     & 0.51               \\ \hline
\textbf{CIFAR10-R56}                                                  & 86.7\%                        & 99.4\%                           & 84.0\%                & 93.9\%                   & 0.41            & 84.5\%                  & 98.9\%                     & 0.41               & 84.0\%                  & 93.9\%                    & 0.41               \\ \hline
\textbf{CIFAR100-R20}                                                 & 46.3\%                        & 75.2\%                        & 45.6\%                & 75.4\%                & 0.63            & 45.6\%                  & 75.4\%                  & 0.63               & 45.6\%                  & 75.4\%                  & 0.63               \\ \hline
\textbf{CIFAR100-R32}                                                 & 44.5\%                      & 74.6\%                        & 41.2\%              & 71.0\%                & 0.52            & 41.2\%                  & 71.0\%                  & 0.52               & 41.2\%                  & 71.0\%                  & 0.52               \\ \hline
\textbf{CIFAR100-R56}                                                 & 45.7\%                        & 75.1\%                        & 44.9\%                & 74.5\%                & 0.68            & 44.9\%                & 74.5\%                  & 0.68               & 44.9\%                  & 74.5\%                  & 0.69               \\ \hline
\textbf{ImageNet200-R20}                                              & 33.2\%                        & 59.7\%                      & 29.6\%                & 56.1\%              & 0.43            & 32.6\%                  & 59.7\%                & 0.75               & 32.6\%                  & 59.7\%                & 0.75               \\ \hline
\textbf{ImageNet200-R32}                                              & 31.8\%                        & 58.6\%                      & 21.0\%                & 58.0\%              & 0.33            & 31.5\%                  & 58.0\%                & 0.72               & 31.5\%                  & 58.0\%                & 0.72               \\ \hline
\textbf{ImageNet200-R56}                                              & 32.1\%                        & 58.7\%                      & 30.0\%                & 56.7\%              & 0.30            & 31.0\%                  & 57.8\%                & 0.71               & 31.0\%                  & 57.8\%                & 0.71               \\ \hline
\end{tabular}

Acc: Accuracy, CI: Concern Identification, TI: Tangling Identification, MC: Modularizing Concern, MC-BLN: Modularizing Concern: Backtrack to Last Convolutional Layer, MC-BI: Modularizing Concern: Backtrack to Input, and JI: Jaccard Index.
\end{table*}

%% file: rq1.tex
\subsection{Results}
In this section, we evaluate our approach to understand the cost involved in decomposition, whether the decomposed modules can be reused or replaced, and how decomposition can be beneficial compared to training from scratch.
\subsubsection{Does Decomposing CNN Model into Module Involve Cost?}
\label{subsec:rq1}
\input{tbintrareuse1.tex}
\input{tbinterreuse1.tex}
To understand how decomposing CNN models into modules performs, we evaluate our approach on 9 datasets and model combinations. First, we decompose the CNN model into small modules, and then we compose them to evaluate how these composed modules perform compared to the trained CNN models. In Table \ref{tb:rq1}, the composed accuracy of the decomposed modules and the trained models' accuracy have been shown. We report the top-1 and top-5 accuracies of the models. Here, in the first and second columns, we show the top-1 and top-5 accuracy of the model. Whereas, in columns 3-11, the accuracy shown is from the composition of the decomposed modules. While composing the modules, we apply the voting-based approach that is similar to the prior work. Also, we compute the Jaccard index (JI) to identify the average variability between the modules and the model. Lesser value of the JI represents better decomposition as the modules are desired to be significantly different from the model. 
Suppose the value of the Jaccard index is very high. In that case, it denotes that the module has essentially become similar to the model.
While the lower JI is a criterion for better decomposition, the cost is another criteria to be considered while decomposing a model into modules. In this study, our objective is to have the least cost of decomposition with the most dissimilarities between the modules and the model.
We found that in all the cases, there is a cost involved while decomposing. For instance, our first approach identifies the concern, adds negative examples, and finally modularizes the concern involves 3.46\% and 2.54\%  (top-1 and top-5) of loss of accuracy with an average Jaccard index 0.50. Whereas applying dense-based backtracking, the loss has reduced. The average loss with this approach is 1.77\% and 0.85\%, and the average Jaccard index is 0.63. For approach involving the backtrack through the convolution layer includes a loss of 1.86\% and 1.93\% accuracy with an average 0.64 Jaccard index. For the further experiments, we choose the second approach as the loss is the least of all. Based on these results, we can conclude that decomposition is possible in CNN models. However, it involves a small cost, and the modules produced are significantly different from the models.
For further studies, we used the dense-based backtracking technique.

%% file: tbintrareuse1.tex
\begin{table*}%
	\centering
	\footnotesize
		\caption{Intra Dataset Reuse. MA: Composed Module Accuracy, TMA: Trained Model Accuracy, C10: CIFAR-10, C100: CIFAR-100, I200: ImageNet-200, and C\{x\}: Output Label x}%
	\subfloat[CIFAR-10 and CIFAR-100 Reuse. \colorbox{moccasin}{Yellow} and \colorbox{magicmint}{Green} represent the CIFAR-10 and CIFAR-100 reuse scenarios, respectively.]{
		
			\begin{tabular}{|c|c|c|c|c|c|c|c|}
				\hline
				\multicolumn{2}{|c|}{\textbf{C10}}           & \multicolumn{2}{c|}{\textbf{C2}}                         & \multicolumn{2}{c|}{\textbf{C3}}                         & \multicolumn{2}{c|}{\textbf{C4}}                         \\ \hline
				\textbf{C100} & \textbf{C10}    & \textbf{TMA}                 & \textbf{MA}                  & \textbf{TMA}                 & \textbf{MA}                  & \textbf{TMA}                 & \textbf{MA}                  \\ \hline
				\hline
				\textbf{C1} & \textbf{C1} & \cellcolor[HTML]{FAEBD7}95.4\% &  \cellcolor[HTML]{FAEBD7}\textbf{98.8\%} &
				 \cellcolor[HTML]{FAEBD7}\textbf{84.7\%}  &
				  \cellcolor[HTML]{FAEBD7}\textbf{91.4\%} &  \cellcolor[HTML]{FAEBD7}\textbf{91.9\%} &
				  \cellcolor[HTML]{FAEBD7}\textbf{96.6\%}  \\ \hline
				\textbf{C2} & \textbf{C2} & \cellcolor[HTML]{AAF0D1}\textbf{91.0\%} & \cellcolor[HTML]{AAF0D1}\textbf{91.0\%} & 
				 \cellcolor[HTML]{FAEBD7}\textbf{93.2\%} &
				 \cellcolor[HTML]{FAEBD7}\textbf{99.3\%} &  \cellcolor[HTML]{FAEBD7}\textbf{92.2\%} &
				 \cellcolor[HTML]{FAEBD7}\textbf{99.3\%} \\ \hline
				\textbf{C3} & \textbf{C3} & \cellcolor[HTML]{AAF0D1}\textbf{96.0\%} & \cellcolor[HTML]{AAF0D1}\textbf{96.0\%} & \cellcolor[HTML]{AAF0D1}\textbf{95.0\%} & \cellcolor[HTML]{AAF0D1}\textbf{98.0\%} & \cellcolor[HTML]{FAEBD7}\textbf{81.2\%} & \cellcolor[HTML]{FAEBD7}\textbf{92.9\%} \\ \hline
				\textbf{C4} & \textbf{C4} & \cellcolor[HTML]{AAF0D1}100\% & \cellcolor[HTML]{AAF0D1}95.5\%   & \cellcolor[HTML]{AAF0D1}100\% & \cellcolor[HTML]{AAF0D1}96\%   & \cellcolor[HTML]{AAF0D1}\textbf{98.5\%}   & \cellcolor[HTML]{AAF0D1}\textbf{100\%} \\ \hline
				\multicolumn{2}{|c|}{\textbf{C100}}          & \multicolumn{2}{c|}{\textbf{C100-C1}}                        & \multicolumn{2}{c|}{\textbf{C100-C2}}                        & \multicolumn{2}{c|}{\textbf{C100-C3}}                        \\ \hline
			\end{tabular}
	}%
	\subfloat[ImageNet-200 Reuse.  \colorbox{moccasin}{Yellow} represents the ImageNet-200 reuse scenarios. \colorbox{black}{\textcolor{white}{Black}} represents the wrong combination of reuses.]{
		
			\begin{tabular}{|c|c|c|c|c|c|c|}
				\hline
				\multirow{2}{*}{\textbf{I200}}
				& \multicolumn{2}{c|}{\textbf{C2}}                        & \multicolumn{2}{c|}{\textbf{C3}}                      & \multicolumn{2}{c|}{\textbf{C4}}                                                        \\ \cline{2-7}
				& \textbf{TMA}                 & \textbf{MA}                  & \textbf{TMA}                & \textbf{MA}                 & \textbf{TMA}                                 & \textbf{MA}                                  \\ \hline
				\textbf{C1}           & \cellcolor[HTML]{FAEBD7}\textbf{90.00\%} & \cellcolor[HTML]{FAEBD7}\textbf{98.00\%} & \cellcolor[HTML]{FAEBD7}\textbf{93.0\%} & \cellcolor[HTML]{FAEBD7}\textbf{94.0\%} & \cellcolor[HTML]{FAEBD7}\textbf{92.0\% }                 & \cellcolor[HTML]{FAEBD7}\textbf{94.0\%}                \\ \hline
				\textbf{C2}           & \cellcolor[HTML]{000000}        & \cellcolor[HTML]{000000}        & \cellcolor[HTML]{FAEBD7}95.0\% & \cellcolor[HTML]{FAEBD7}88.0\% & \cellcolor[HTML]{FAEBD7}\textbf{95.0\% }                 & \cellcolor[HTML]{FAEBD7}\textbf{95.0\%}                  \\ \hline
				\textbf{C3}           & \cellcolor[HTML]{000000}        & \cellcolor[HTML]{000000}        & \cellcolor[HTML]{000000}       & \cellcolor[HTML]{000000}       & \cellcolor[HTML]{FAEBD7}\textbf{89.0\%}                  & \cellcolor[HTML]{FAEBD7}\textbf{94.0\% }                 \\ \hline
				\textbf{C4}           & \cellcolor[HTML]{000000}        & \cellcolor[HTML]{000000}        & \cellcolor[HTML]{000000}       & \cellcolor[HTML]{000000}       & \cellcolor[HTML]{000000}{\color[HTML]{333333} } & \cellcolor[HTML]{000000}{\color[HTML]{333333} } \\ \hline
			\end{tabular}
	}

	\label{tb:reuse}%
\end{table*}

%
%
%

%% file: tbinterreuse1.tex
\begin{table*}[htp]
	\caption{Inter Dataset Reuse.}
	\centering
	\footnotesize
\begin{tabular}{|l|r|r|r|r|r|r|r|r|r|r|r|r|}
	\hline
	\multirow{2}{*}{\backslashbox{\textbf{C100}}{\textbf{C10}}}& \multicolumn{3}{c|}{\textbf{C1}} & \multicolumn{3}{c|}{\textbf{C2}} & \multicolumn{3}{c|}{\textbf{C3}} & \multicolumn{3}{c|}{\textbf{C4}} \\
	\cline{2-13}
	& \multicolumn{1}{|l}{\textbf{TMA}} & \multicolumn{1}{l|}{\textbf{MA}} & \multicolumn{1}{l|}{\textbf{MMA}} & \multicolumn{1}{l|}{\textbf{TMA}} & \multicolumn{1}{l|}{\textbf{MA}} & \multicolumn{1}{l|}{\textbf{MMA}} & \multicolumn{1}{l|}{\textbf{TMA}} & \multicolumn{1}{l|}{\textbf{MA}} & \multicolumn{1}{l|}{\textbf{MMA}} & \multicolumn{1}{l|}{\textbf{TMA}} & \multicolumn{1}{l|}{\textbf{MA}} & \multicolumn{1}{l|}{\textbf{MMA}} \\
	\hline
	\hline
	\textbf{C1} & 97.8\% & 83.0\% & 91.5\% & 96.7\% & 87.9\% & 92.1\% & 96.6\% & 86.0\% & 91.5\% & 98.8\% & 87.2\% & 93.8\% \\
	\hline
	\textbf{C2} & 95.5\% & 82.4\% & 89.2\% & 98.6\% & 87.5\% & 88.1\% & 95.6\% & 85.2\% & 89.9\% & 99.7\% & 84.1\% & 89.0\% \\
	\hline
	\textbf{C3} & 98.4\% & 78.2\% & 87.5\% & 97.9\% & 79.5\% & 88.5\% & 94.6\% & 80.2\% & 89.2\% & 99.6\% & 81.6\% & 87.9\% \\
	\hline
	\textbf{C4} & 98.0\% & 88.3\% & 88.1\% & 97.2\% & 88.6\% & 88.1\% & 94.7\% & 86.7\% & 90.3\% & 99.0\% & 88.9\% & 91.9\% \\
	\hline
\end{tabular}\\
TMA: Trained Model Accuracy, MA: Composed Module Accuracy, and MMA: Modified Module Accuracy.
\label{tb:interreuse}
\end{table*}

%% file: rq2.tex
\subsubsection{How module reuse and replacement can be done compared to retraining a model from scratch?} Here, we evaluate how decomposed modules can be reused and replaced in various scenarios.
\label{subsec:rq2}
\input{tbintrareplace.tex}
\input{tbinterreplace.tex}

\textbf{Reusability:} Table \ref{tb:reuse}, and \ref{tb:interreuse} show two different reuse scenarios: 1) intra-dataset reuse, 2) inter-dataset reuse.

In intra-dataset reuse, we build a model by taking two output classes from a dataset and building a new model. For instance, output classes 0 and 1 have been taken from CIFAR-10, and we evaluate the accuracy by combining the decomposed modules for output classes 0 and 1 (we represent the output class with their index in the dataset). We compare the performance with retraining a model with the same structure with the same output classes and measure the accuracy. We took the best model for each dataset (based on the trained model accuracy) and the modules created from that. For CIFAR-10, CIFAR-100, and ImageNet-200, the best model in terms of training accuracy are ResNet-56, ResNet-20, and ResNet-20, respectively. Since these models take a long time to train, we performed the experiments for 4 output classes randomly chosen from the dataset to evaluate the reusability scenarios. In Table \ref{tb:reuse}, the intra-dataset reuse scenarios are reported. C1, C2, C3, and C4 denote the four randomly chosen classes, and they are different for each dataset.
Since we take 2 output classes in each experiment, the total number of choices for a dataset with $n$ output classes under experiment is $n \choose 2$. In our cases, we take 4 output classes from each dataset to evaluate, and the total choices would be $4 \choose 2$ or 6. For CIFAR-10, reusing the modules increase 6.6\% accuracy, on average. For CIFAR-100, there is an average 0.67\% loss, but 4/6 cases reusing the modules perform better than the trained model. For ImageNet, there is an average 1.5\% increase in accuracy. Overall, the gain in the accuracy in intra-dataset reuse is 2.5\%. 

In inter-dataset reuse, we take two output classes from different datasets and build a binary classification-based problem. For example, output classes 1 and 2 have been taken from the CIFAR-10 and CIFAR-100 datasets, respectively. We compute the accuracy the same way as we did for intra-dataset reuse scenario. Since the model structure of the decomposed modules for two datasets are different, e.g., the decomposed modules for CIFAR-10 follow the model structure of ResNet-56, whereas it is ResNet-20 for CIFAR-100. The best model in terms of accuracy has been taken for retraining. We did a pilot study and found that ResNet-20 does better in terms of accuracy for the inter-dataset reuse scenario for CIFAR-10 and CIFAR-100 datasets. To overcome overfitting, we store the checkpoints for all training-based evaluations and report the last checkpoint with the best validation accuracy. In Table \ref{tb:interreuse}, the inter-dataset reuse scenarios are reported. We found that inter-dataset reuse has a non-trivial cost. In this case, the loss of accuracy is 12.06\%. 
Our intuition is that since these modules originate from a different dataset, they still have some traits of the parent dataset. In fact, by applying the tangling identification approach, we deliberately add some non-concerned examples to distinguish between the concerned and the non-concerned examples. 
To alleviate the effect, we perform a continuous learning-based approach, where we re-apply the tangling identification with the unseen output class(es). 
We found that enabling continuous learning can reduce the cost of decomposition significantly. The overall loss while reusing modules is 7.63\%, which is a 4.4\% gain compared to the previous reusability approach. Since the input size of the CIFAR-10 and CIFAR-100 images are not the same as ImageNet-200, both training a single model and reusing modules cannot be done. 

\textbf{Replaceability:}
Similar to the reusability scenario, we evaluated both inter and intra-dataset reuses. For intra dataset reuse, we replace a module decomposed from a model with less accuracy with a module for the same output class decomposed from a model with higher accuracy. In Table \ref{tb:intrareplace}, we report the evaluation for replacing the module. For each dataset, we evaluate for 10 output classes to match the total number of classes for all the datasets. For CIFAR-10, we replace a module decomposed from ResNet-32 with a module for the same output class decomposed from ResNet-20. We found that the intra-dataset reuse increases the accuracy by 0.7\% and 1.0\% for top-1 and top-5 accuracy, respectively in comparison to the prior composed accuracy of the decomposed modules. In fact, 80.0\% (24/30) and 83.3\% (25/30) times replacing a module does the better or the same compared to the composed accuracy of the decomposed modules for top-1 and top-5 accuracy, respectively. Also, for 33.3\% and 66.7\% cases, reusing modules does better than the model accuracy for top-1 and top-5 accuracies, respectively.
Although we incorporated the continuous learning approach, there is no significant increase in accuracy  (top-1: 0.02\%  and top-5: 0.01\%).

Inter-dataset replacement scenarios are reported in Table \ref{tb:interreuse}. To reduce the experiment time, we replace a module from CIFAR-10 with a module taken randomly from CIFAR-100 and report both composed accuracy and the accuracy of the trained model for four different output classes. We found that for top-1 accuracy, there is a 2.5\% accuracy increase, and for top-5, there is a loss of 7.1\%, on average.
\input{motivationtb.tex}

\noindent\textbf{Motivating Example.} Here, we recreate the example shown in Figure \ref{fig:overview}. Since human face and Gorilla images are not present in the ImageNet-200 dataset, we added two classes from the ImageNet large dataset. However, there is no class specifically categorized as ``human face'' in the dataset. So, we take the closest class that contains the images of different persons.
First, we build a model with 202 (ImageNet-200 + Person + Gorilla) output classes. For the first option, we create a hypothetical model that does not exhibit faulty behavior. We do that by training a model with more epochs (+100 epochs) and achieve higher accuracy (+1.8\%). Then, we decompose the model into modules. Then, we replace the Gorilla module with the module created from the second model. In the second option, we train a model with Person and Gorilla examples and replace the faulty Gorilla module with the newly created one. In the third option, we remove the Gorilla module from the set of modules. Finally, in the fourth option, if the original model predicts Person or Gorilla, we re-verify with the modules created from the 2-class classification model in the second option. The evaluation has been shown in Table \ref{tb:motivating}, and we found that enabling the replacement and reuse of modules can solve the problem. Based on the need and available resources, users can pick any of the four options.

%% file: tbinterreplace.tex
\begin{table*}[]
	\caption{Intra Dataset Replace.}
	\footnotesize
	\centering
	\label{tb:intrareplace}
	\vspace{-10pt}
	\begin{tabular}{|l|c|c|c|c|c|c|c|c|c|c|c|c|}
		\multicolumn{13}{c}{\textbf{Top-1 Accuracy}}\\
		\hline
		\textbf{Dataset} & \textbf{TMA} & \textbf{Prior MA} &\textbf{RM0} & \textbf{RM1} &\textbf{RM2} & \textbf{RM3} & \textbf{RM4} & \textbf{RM5} & \textbf{RM6} & \textbf{RM7} & \textbf{RM8} & \textbf{RM9} \\
		\hline
		\hline
		\textbf{CIFAR-10} & 85.3\% & 78.6\% & 80.6\% & 79.2\% & 80.0\% & 80.9\% & 80.7\% & 80.2\% & 79.8\% & 79.6\%& 79.4\%& 79.6\% \\
		\hline
		\textbf{CIFAR-100} & 44.5\% & 41.2\% & 40.7\% & 41.0\% & 41.2\%& 41.1\% & 41.2\% & 41.1\% & 41.2\% & 41.2\% & 41.1\% & 41.0\% \\
		\hline
		\textbf{ImageNet-200} & 31.8\% & 31.5\% & \textbf{ 32.4\% }   &    \textbf{ 32.2\%}&       \textbf{32.3\%}&       \textbf{32.0\%}&       \textbf{32.3\%}&       \textbf{32.4\%}&       \textbf{32.4\%}&      \textbf{33.3\%}&       \textbf{33.2\%}&  \textbf{33.3\%}\\
		\hline
		\multicolumn{13}{c}{\textbf{Top-5 Accuracy}}\\
		\hline
		\textbf{Dataset} & \textbf{TMA} & \textbf{Prior MA} &\textbf{RM0} & \textbf{RM1} &\textbf{RM2} & \textbf{RM3} & \textbf{RM4} & \textbf{RM5} & \textbf{RM6} & \textbf{RM7} & \textbf{RM8} & \textbf{RM9} \\
		\hline
		\hline
		\textbf{CIFAR-10} & 99.2\% & 98.1\% & \textbf{100\%} & \textbf{100\%} & \textbf{100\%} & \textbf{100\%} & \textbf{100\%} & \textbf{100\%} & \textbf{100\%} & \textbf{100\%} & \textbf{100\%} & \textbf{100\% }\\
		\hline
		\textbf{CIFAR-100} & 74.6\% & 71.0\% & 70.7\% & 70.8\% & 71.0\% & 70.9\% & 71.0\% & 71.0\% & 71.1\% & 71.1\% & 70.9\% & 70.9\% \\
		\hline
		\textbf{ImageNet-200} & 58.6\% & 58.0\% &  \textbf{59.3\%}     & \textbf{59.1\%}  &       \textbf{59.1\%}&       \textbf{59.2\%}&       \textbf{59.0\%}&       \textbf{59.2\%}&       \textbf{59.1\%}&     \textbf{59.2\%}  &      \textbf{59.2\%}&\textbf{59.1\%}\\
		\hline
	\end{tabular}%
	
	TMA: Trained Model Accuracy, MA: Composed Module Accuracy, RM\{x\}: Replace Module x with another Module for x.
\end{table*}
\begin{table*}[]
	\caption{Inter Dataset Replace. (All results are in \%.)}
	\footnotesize
	\centering
     \begin{tabular}{|c|r|r|r|r|r|r|r|r|r|r|r|r|r|r|r|r|}
     	\hline
    \multirow{2}{*}{\backslashbox{\textbf{C100}}{\textbf{C10}}}& \multicolumn{4}{c|}{\textbf{C1}}         & \multicolumn{4}{c|}{\textbf{C2}}         & \multicolumn{4}{c|}{\textbf{C3}}         & \multicolumn{4}{c|}{\textbf{C4}} \\
    	\cline{2-17}
    	& \textbf{M1} & \textbf{D1} & \textbf{M5} & \textbf{D5} & \textbf{M1} & \textbf{D1} & \textbf{M5} & \textbf{D5} & \textbf{M1} & \textbf{D1} & \textbf{M5} & \textbf{D5} & \textbf{M1} & \textbf{D1} & \textbf{M5} & \textbf{D5} \\
    	\hline
    	\hline
    	\textbf{C1}     & 86.2\ & \textbf{87.3}\ & 99.6\ & 95.9\ & 83.1\ & \textbf{87.4}\ & 98.9\ & 95.6\ & 86.1\ & \textbf{87.1}\ & 99.5\ & 95.4\ & 86.7\ & \textbf{87.0}\ & 99.6\ & 95.4\ \\
    	\hline
    	\textbf{C2}     & 85.3\ & \textbf{86.3}\ & 99.2\ & 90.8\ & 82.7\ & \textbf{86.3}\ & 98.9\ & 90.5\ & 84.6\ & \textbf{86.1}\ & 99.1\ & 90.3\ & 87.6\ & 86.0\ & 99.4\ & 90.3\ \\
    	\hline
    	\textbf{C3}     & 86.2\ & \textbf{88.1}\ & 99.4\ & 91.5\ & 81.8\ & \textbf{88.3}\ & 98.8\ & 91.3\ & 86.2\ & \textbf{88.0}\ & 99.0\ & 91.2\ & 87.9\ & 87.8\ & 99.6\ & 91.3\ \\
    	\hline
    	\textbf{C4}     & 84.7\ & \textbf{89.9}\ & 99.2\ & 91.4\ & 82.6\ &\textbf{89.7}\ & 98.7\ & 91.1\ & 85.0\ & \textbf{89.5}\ & 98.8\ & 91.1\ & 86.6\ & \textbf{89.3}\ & 99.4\ & 91.1\ \\
    	\hline
    \end{tabular}

M1 and  M5: Top-1 and Top-5 Accuracy for Trained Model D1 and D5: Top-1 and Top-5 Composed Accuracy for Decomposed Modules.
\end{table*}

%% file: motivationtb.tex
\begin{table}[htp]
	\footnotesize
	\centering
	\caption{Scenarios Created Based on Figure \ref{fig:overview}.}
	\label{tb:motivating}
\begin{tabular}{|l|c|c|}
	\hline
\textbf{Strategies}                                                                      & \textbf{Top-1 Acc}                  & \textbf{Top-5 Acc}                  \\
\hline
\hline
Initial Model                                                                   & 26.6\% &50.4\% \\
\hline
MA& 26.0\% & 50.4\% \\
\hline
Option 1                                                                        & 27.2\% & 54.6\% \\
\hline
Option 2                                                                        &                   25.0\%         &           50.4\%                 \\
\hline
Option 3                                                                        &            24.0\%                &   48.2\%  \\
\hline        
Option 4                                                                        &            25.3\%                &   50.8\%  \\
\hline                 
\end{tabular}
\end{table}

%% file: rq3.tex
\subsubsection{Does reusing and replacing the decomposed modules emit less $CO_2e$?}
\label{subsec:rq3}
\begin{figure}[!ht]
	\centering
	\includegraphics[, trim={2cm 7cm 0cm 7cm}, width=1.12\linewidth]{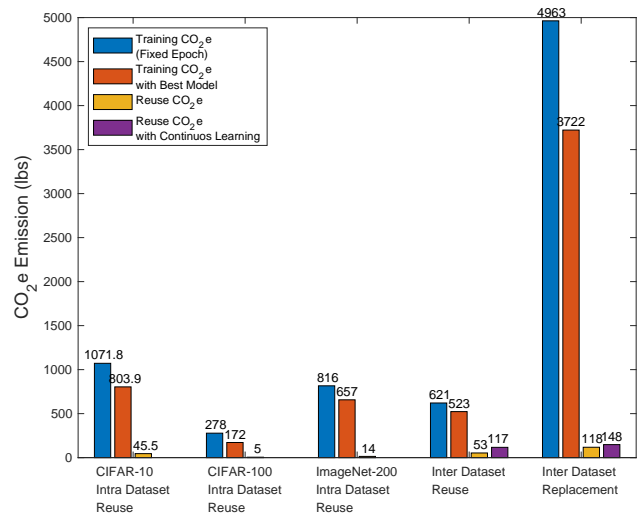}
	\caption{Comparison of $CO_2e$ Emission.}
	\label{fig:c02}
\end{figure}
Strubell~\etal~\cite{strubell2019energy} have identified that training a model often emits six times more $CO_2e$ than a car emits over a year. To understand how decomposition can help to reduce the harmful effect, we measure the $CO_2e$ emission for both intra and inter-dataset reuse and replace scenarios. First, we start computing the power consumption before executing the program. Then, we measure the average power drawn by other tasks running in the background for 10 sec and compute the average power drawn by CPU and DRAM. We negate these two values to separate the resource consumption of the program and other background tasks. Then, we measure the power consumption for each 100 ms (default for Intel Power Gadget). 
Figure \ref{fig:c02} shows the $CO_2e$ emission for different scenarios. We do not show intra dataset replacement scenarios as that cannot be compared with training from scratch.
For training scenarios, we build the model from scratch, and after training, we predict the same set of images. This experimental setting has been done to compare a similar situation, where developers need to build and predict inputs. The value reported in the figure is the average $CO_2e$ consumption for all the experimented scenarios described in \S\ref{subsec:rq2}. Also, since the epoch has been fixed for each retraining, the power consumption is somewhat fixed for each retraining scenario. However, the best model can be found earlier. For instance, a model is trained with 100 epochs, but the best model is found at the 20th epoch. To remove the effect of overfitting, we compute the power consumed until the return of the best model and report that in the figure. 
We found that for reuse scenarios, decomposition reduces the $CO_2e$ consumption by 
23.7x and 18.3x for the fixed epoch and the best model scenarios, respectively. For replacement scenarios, it is 42x and 31.5x, respectively. If we apply the continuous learning-based approach, there is a slight increase in resource consumption, but still significantly less than training from scratch. Also, we computed the additional $CO_2e$ consumption for the one-time decomposition approach and found that on average 116, 371, and 3800 lbs of $CO_2e$ has been generated for decomposing CIFAR-10, CIFAR-100, and ImageNet-200 models, respectively. The overhead is significantly lower for CIFAR-10 and CIFAR-100 compared to training a new model for both reuse and replace scenarios. However, for ImageNet-200, the overhead is high, but it is a one-time operation that can enable both reuses and replacements at a very low cost.

%% file: conclusion.tex
\section{Conclusion and Future Work}
\label{sec:conclusion}
In this paper, we introduce decomposition in convolutional networks and transform a trained model into smaller components or modules. We used a data-driven approach to identify the section in the model responsible for a single output class and convert that into a binary classifier. 
Modules created from the same or different datasets can be reused or replaced in any scenario if the input size of the modules is the same. We found that decomposition involves a small cost of accuracy. However, both intra-dataset reuse and replaceability increase the accuracy compared to the trained model. Furthermore, enabling reusability and replaceability reduces $CO_2e$ emission significantly. For this work, we omit the heterogeneous inputs (the input size for modules are not the same) while reusing and replacing modules, and it will be a good 
research direction for the future to study how an interface could be built around the modules to take different types of inputs.